\documentclass[runningheads]{llncs}

\usepackage{eccv}

\usepackage{eccvabbrv}

\usepackage{graphicx}
\usepackage{booktabs}

\usepackage[accsupp]{axessibility}  

\usepackage{hyperref}

\usepackage{orcidlink}

\usepackage{enumitem}
\usepackage{subcaption}

\begin{document}

\title{To Supervise or Not to Supervise: Understanding and Addressing
the Key Challenges of Point Cloud Transfer Learning} 

\titlerunning{To Supervise or Not to Supervise}

\author{Souhail Hadgi\inst{1} \and
Lei Li\inst{2} \and
Maks Ovsjanikov\inst{1}}

\authorrunning{S.~Hadgi et al.}

\institute{LIX, Ecole Polytechnique, IP Paris \and
           Technical University of Munich}
\newcommand{\mypara}[1]{\noindent\textbf{#1}~}
\newcommand{\itpara}[1]{\noindent\textit{#1}~}
\maketitle

\begin{abstract}
    Transfer learning has long been a key factor in the advancement of many fields including 2D image analysis. Unfortunately, its applicability in 3D data processing has been relatively limited. While several approaches for point cloud transfer learning have been proposed in recent literature, with contrastive learning gaining particular prominence, most existing methods in this domain have only been studied and evaluated in limited scenarios. Most importantly, there is currently a lack of principled understanding of both $\textit{when}$ and $\textit{why}$ point cloud transfer learning methods are applicable. Remarkably, even the applicability of standard \textit{supervised} pre-training is poorly understood. In this work, we conduct the first in-depth quantitative and qualitative investigation of supervised and contrastive pre-training strategies and their utility in downstream 3D tasks. We demonstrate that layer-wise analysis of learned features provides significant insight into the downstream utility of trained networks. Informed by this analysis, we propose a simple geometric regularization strategy, which improves the transferability of supervised pre-training. Our work thus sheds light onto both the specific challenges of point cloud transfer learning, as well as strategies to overcome them.
\end{abstract}
\section{Introduction}
\label{sec:Introduction}

Transfer learning is an integral part of the success of deep learning in 2D computer vision, enabling the use of powerful pre-trained architectures, which can be adapted with limited data on a broad range of downstream tasks and datasets \cite{he2016deep,chen2020simple,zhao2021what}. The advancement of transfer learning can be attributed to several factors such as the availability of well-established architectures (\eg{} deep CNNs \cite{he2016deep}, ViTs \cite{dosovitskiy2020image}), an abundance of source data to train on (\eg{} ImageNet \cite{deng2009imagenet} among myriad others), as well as the development of different pre-training strategies \cite{jaiswal2020survey}.

\begin{figure}[tp]
    \hspace*{-10mm}
    \centering    \includegraphics[width=0.95\linewidth]{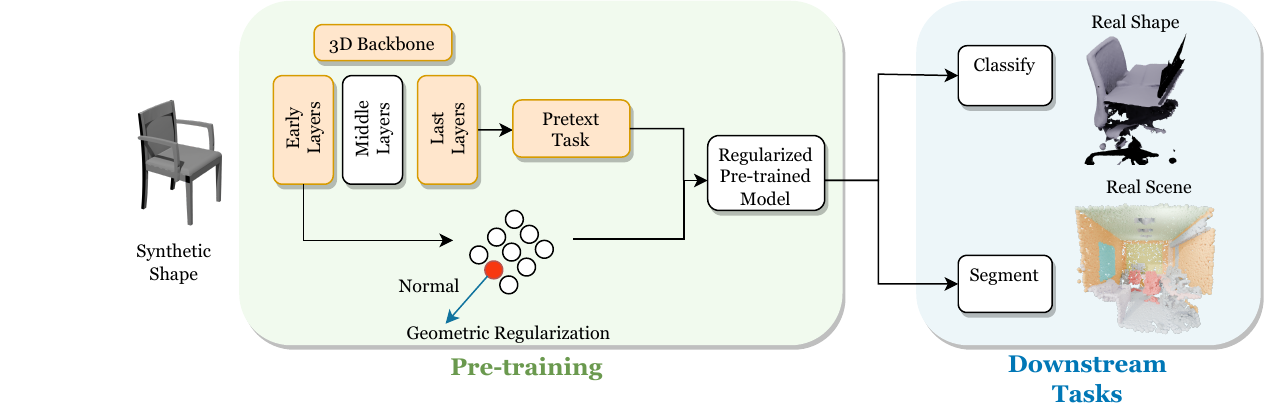}
    \caption{\textbf{Analyzing and improving point cloud transfer learning.} In this work, we perform the first in-depth study of the different components of network pre-training that influence the outcome of point cloud transfer learning (components in solid orange). This includes the source data domain in relation to the downstream target data, the choice of architecture, the importance of early vs later layers, and the pre-training design choices. We also show that improvements to the pipeline can be achieved through regularization of the early layers, by promoting the prediction of \textit{geometric} properties.}
    \label{fig:rendering}
\end{figure}

Unfortunately, the extension of such approaches to the 3D domain presents several important challenges. First, and perhaps most importantly, large-scale labeled datasets are scarce in 3D due to the complexity of data acquisition and annotation as well as significant domain-specificity that creates imbalances across data (\eg{} synthetic CAD shapes \cite{shapenet2015,wu20153d,koch2019abc}, real scenes \cite{Dai_2017_CVPR_scannet}, non-rigid shapes \cite{bogo2017dynamic}, etc.). Such imbalances, coupled with limited training data, can strongly hinder the applicability of transfer learning solutions \cite{xie2020pointcontrast}.

Another key challenge in learning with 3D data is the  nature of the data \textit{representation}, which most often comes in the form of unstructured point clouds, potentially with attributes (\eg{} spatial, color). This limits the use of architectures that rely on regular grid structure, and furthermore, poses challenges due to significant variability in \textit{sampling} properties (density, acquisition artefacts, etc.). As a result, while a large number of different architectures have been proposed for 3D deep learning \cite{guo2020deep}, there is a lack of universal solutions applicable to an arbitrary scenario.

In spite of these challenges, several works have recently emerged aiming to enable transfer learning for point clouds, especially by focusing on new pre-training strategies, \eg{}, \cite{xie2020pointcontrast,hou2021exploring,xiao2022unsupervised}. Among these strategies, several approaches have shown that \textit{self-supervised strategies} (\eg{} contrastive learning, point cloud reconstruction) can improve performance on different downstream data (\eg{} shapes, scenes) and on select set of tasks and datasets.

Despite the growing number of these approaches, unfortunately, most existing methods are only presented and evaluated in very specific settings, and missing comparison to baselines such as supervised pre-training. Specifically, there are several questions not addressed in existing literature, including: 1) The utility of supervised compared to contrastive pre-training, 2) The role of the architecture's inherent properties in the success of point cloud transfer learning, and 3) The impact of regularization strategies on the efficacy of point cloud pre-training. 

By performing the first in-depth investigation of the most prominent approaches, we observe that in the \textit{linear probing setting}, supervised pre-training leads to superior performance compared to contrastive learning across almost all tested architectures, even under relatively significant domain shift. This suggests that within 3D deep learning, contrastive learning does not always lead to more ``universally useful'' features. At the same time, we also observe that in the full \textit{fine-tuning scenario}, contrastive pre-training can outperform supervised learning. By analyzing gradients of pre-trained networks, we show that the \textit{early layers} learned with supervised pre-training are not easily adaptable across new datasets, compared to those learned with contrastive learning. This sheds new and specific light on the differences between these pre-training strategies in the context of point cloud analysis. Finally, we demonstrate that a simple regularization scheme applied to the early layers during supervised pre-training can overcome this limitation and leads to networks that can be fine-tuned efficiently.

To summarize, our main contributions are as follows:
\begin{enumerate}[topsep=1pt,itemsep=0ex,partopsep=1ex,parsep=1ex]
\item We perform the first comparison, within a single consistent evaluation framework, of supervised and contrastive pre-trained models using different point cloud backbones, assessing their transfer learning performance on common downstream data and tasks. We separate between the linear probing and fine-tuning settings and analyze the difference in performance.
\item We observe a clear quantifiable difference in architecture's layer properties and find that, remarkably, for 3D architectures, \textit{even the first layers} have the ability to discriminate between downstream shape classes. Moreover we show that supervised pre-training can lead to early layers that do not easily adapt to new data.

\item To counteract this effect, we introduce a simple \textit{layer-wise} geometric regularization procedure on supervised pre-training, which leads to performance improvement and outperforms contrastive learning in several settings. We finally strongly correlate the improvement to the key signals identified in layer gradient norm. 
\end{enumerate}

\noindent Throughout all our analysis, we take special care to make all models and evaluation settings consistent and comparable. 

\section{Related Work}
\label{sec:RelatedWork}
Transfer learning has facilitated myriad applications in 2D image analysis, such as segmentation \cite{Long_2015_CVPR,he2017mask}, detection \cite{ren2015faster}, style transfer \cite{gatys2016image}, and medical image analysis \cite{tajbakhsh2016convolutional,Matsoukas_2022_CVPR}, among others.
This can be largely attributed to the success of deep neural networks \cite{AlexNet:2012,he2016deep} on the large-scale visual recognition dataset ImageNet \cite{deng2009imagenet}.
In contrast, research on generic 3D representation learning, driven by the development of 3D deep learning techniques \cite{qi2017pointnet,wang2019dynamic,choy20194d} in recent years, is still emerging.

Contrastive learning has garnered growing attention recently as a promising unsupervised representation learning paradigm, which can be conceptualized as a dictionary look-up process \cite{he2020momentum}.
In this process, a query and a set of keys are encoded by some network, and a contrastive loss seeks to maximize the similarity between the query and its single positive key and minimize that between the query and its all negative keys.

For contrastive pre-training in 3D, several studies have contributed to the design of effective learning setups.
In particular, Xie \etal{} introduced PointContrast \cite{xie2020pointcontrast}, which utilizes two rigidly transformed views of point clouds and performs point-level feature contrasting in the two views.
They demonstrated that the pre-trained network brings noticeable improvement in downstream 3D learning tasks, including object classification~\cite{shapenet2015}, part segmentation~\cite{yi2016scalable}, semantic segmentation~\cite{armeni20163d,Dai_2017_CVPR_scannet}, and object detection~\cite{Song_2015_CVPR_sunrgbd,Dai_2017_CVPR_scannet}.
Follow-up works investigated this direction further by constraining point contrasting in local regions \cite{hou2021exploring}, adopting patch-level \cite{du2021self}, object-level \cite{rao2021randomrooms}, object and point-level, or even scene-level \cite{zhang2021self,huang2021spatio} contrasting, as well as incorporating image data into the 3D feature contrasting via point-pixel pairs~\cite{liu2020p4contrast,liu2021learning}.
However, most of the aforementioned works focus on demonstrating effective representation learning with specific network architectures and data modalities, lacking a systematic study of both when and why contrastive pre-training enables 3D transfer learning.
In response to this gap, our work delves into the contrastive pre-training strategies across various backbones and downstream tasks, aiming to reveal the essential aspects for a successful 3D transfer learning.

To achieve self-supervised pre-training, several studies have examined an alternative reconstruction-based learning strategy, such as recovering missing points~\cite{wang2021occo,yu2021point}, solving 3D jigsaw tasks~\cite{sauder2019self,sharma2020self,eckart2021self,chen2021shape,alliegro2021joint}, or augmenting auto-encoding with clustering and classification tasks~\cite{hassani2019unsupervised}.
We refer the interested reader to \cite{sun2021adversarially} for a more comprehensive study of this pre-training paradigm.

Finally, we note that several works have aimed to analyze the factors behind successful transfer learning approaches in 2D image tasks \cite{raghu2019transfusion,zhao2021what,Matsoukas_2022_CVPR}. These works point to the special importance of early and mid-level features and to layer-wise analysis of learned representations. Such analysis has not yet been performed for 3D data, and our main objective is to fill this gap by performing a first comprehensive investigation of this topic. As we mention below, our work highlights the special nature of learning on 3D data, thus shedding light on the challenges of point cloud transfer learning and possible ways to address them.

\section{Datasets and Architectures}
\label{sec:DatasetsModels}

\subsection{Datasets}
\mypara{Pre-training Dataset.}
Prior 3D transfer learning approaches \cite{wang2021occo,afham2022crosspoint,xie2020pointcontrast,hou2021exploring} have typically focused on pre-training datasets with limited domain shifts to downstream tasks, such as within the synthetic shape setting \cite{shapenet2015,wu20153d} or the real scene scenario \cite{Dai_2017_CVPR_scannet,armeni20163d}.
        Interestingly, among them, PointContrast \cite{xie2020pointcontrast} argued against pre-training on synthetic data and stated that supervised pre-training is an upper bound to information gain from pre-training (see Sec.~3.1 therein). Our experimental findings (\cref{sec:Evaluation}), however, suggest that this is not always the case and  help to refine this statement.   We thus select ShapeNet \cite{shapenet2015} to evaluate the utility of pre-training with synthetic data, and focus on its transfer utility in downstream tasks with varying domain shifts, as elaborated below.

\mypara{Downstream Datasets and Tasks.}
To comprehensively investigate different 3D transfer learning scenarios, we consider downstream tasks encompassing both 1) \emph{synthetic} vs. \emph{real} data and 2) \emph{object} vs. \emph{scene}-level 3D data, which exhibit a wide range of similarity to ShapeNet (51,300 annotated 3D synthetic shapes in 55 categories),  used for pre-training:

\begin{itemize}
    \item \textbf{ModelNet40} \cite{wu20153d} is a synthetic object-level dataset with 3D shapes resembling those in ShapeNet. It consists of 12,311 shapes and 40 object classes for the shape classification task, and 20 of these classes can be found in ShapeNet.  

    \item \textbf{ScanObjectNN} \cite{uy2019revisiting} is a real object-level dataset with scanned 3D shapes featuring noise and background, vastly different from the clean shapes in ShapeNet. There are 15,000 shapes and 15 object classes for the shape classification task, and 9 of these classes can be found in ShapeNet. 

    \item \textbf{S3DIS} \cite{armeni20163d} is a real scene-level dataset consisting of 3D scans of 6 large-scale indoor areas in office buildings and 13 semantic categories for the semantic segmentation task. 

\end{itemize}

\subsection{Architectures}
To provide an extensive study that tackles diverse 3D backbones, we follow a broadly adopted categorization of 3D deep neural networks \cite{xiao2023unsupervised} and select a diverse set of five representative architectures as follows:

\begin{itemize}
    \item \textbf{Point-based}. PointNet \cite{qi2017pointnet} is a pioneering 3D backbone that is characterized by its \textit{global} nature in feature learning, which has been shown to be beneficial in robust point cloud analysis \cite{taghanaki2020robustpointset}, but, as we show below, can overfit to pre-training data.
    In addition, we consider PointMLP \cite{ma2022rethinking}, which is a recent extension of PointNet and its hierarchical version of PointNet++ \cite{qi2017pointnet++}, and which effectively incorporates local geometry and residual MLPs.
    \item \textbf{Graph-based}. DGCNN \cite{wang2019dynamic} leverages graph convolution operations with graphs constructed in the feature space, allowing it to capture both local and global properties. This architecture has been shown to generalize well between real and synthetic data \cite{uy2019revisiting} and has also demonstrated strong transfer learning performance across various pre-training strategies \cite{wang2021occo,sauder2019self,afham2022crosspoint}. 
    \item \textbf{Sparse convolutions}. MinkowskiNet \cite{choy20194d} is built upon generalized sparse convolutions, which operate on voxels and benefit from the sparsity of point clouds. This architecture is the backbone of PointContrast \cite{xie2020pointcontrast} for 3D transfer learning. Below we compare it to other baselines and examine how its reliance on density impacts its transfer learning capabilities.
    \item \textbf{Transformer}. Following the success of transformers in NLP and 2D vision tasks, 3D transformer backbones have emerged for point cloud processing. For the first time, we study the transfer learning capabilities of Point Cloud Transformer (PCT) \cite{guo2021pct} in downstream tasks. 
\end{itemize}

Our primary objectives in analyzing these architectures are three-fold: firstly, to carry out a first comprehensive analysis of supervised pre-training performance across diverse architectures. Secondly, we explore the relation between architectural properties and their performance in 3D transfer learning, considering both linear probing and fine-tuning scenarios. Thirdly, we aim to identify shared properties that either hinder or contribute to successful transfer learning.

We adapt the layer configurations of these architectures to ensure consistency across all explored pre-training strategies (\cref{sec:Pre-trainingApproaches}).
Further details on the architecture configurations are provided in the supplementary materials.
\section{Pre-training Approaches}
\label{sec:Pre-trainingApproaches}

As mentioned above, in our analysis, we focus on comparing supervised and contrastive learning approaches. Although our primary focus is on these pre-training strategies, we include additional results to alternative methods (including shape-level contrastive learning and a reconstruction-based approach PointDAE \cite{zhang2022point}) in the supplementary materials.

\textbf{Supervised pre-training} is a standard approach used in 2D computer vision tasks to obtain generalizable feature embeddings. Interestingly, its utility in the context of 3D transfer learning has not been extensively investigated in previous studies with a few exceptions such as PointContrast~\cite{xie2020pointcontrast} as mentioned above. In training, we use the standard cross-entropy loss as a pretext objective:
\begin{equation}
    L_{ce} = -\sum_{i}(y_i \cdot \log(p_i)),
\end{equation}
where $y_i$ is the ground truth label, and $p_i$ is the predicted probability of each class $i$ for an input shape. We also use data augmentation (translation, scale and rotation) to avoid the orientation bias introduced by the oriented shapes of ShapeNet, which are also present in the ModelNet40 downstream data. 

\textbf{Contrastive pre-training} is employed for comparisons with supervised pre-training. We use \emph{point-level} contrastive learning, based on PointContrast~\cite{xie2020pointcontrast}.
Specifically, for contrastive pre-training, we first introduce a set of invariances by applying data augmentation. We start by normalizing a point cloud and perform random geometric transformations, including translation with magnitude of 0.5, scaling between 80\% and 125\%, and rotation of magnitude 45$^{\circ}$. We also simulate partial data by cropping 50\% the point cloud.
 
More details about transformations can be found in the supplementary materials. By combining these transformations, we start from an input point cloud and generate two augmented views $P_i$ and $P_j.$ 

For point-level contrastive learning, we contrast between \textit{points} rather than entire objects. This models point-level information and also naturally leads to more data for constructing positive and negative pairs. Positive pairs are obtained by matching points from different views. We use the PointInfoNCE Loss, introduced in \cite{xie2020pointcontrast} as the pretext objective:
\begin{equation}
\mathcal{L}_{Contrastive}=-\sum_{(i, j) \in \mathcal{P}} \log \frac{\exp \left(\mathbf{h}_i \cdot \mathbf{h}_j / \tau\right)}{\sum_{(\cdot, k) \in \mathcal{P}} \exp \left(\mathbf{h}_i \cdot \mathbf{h}_k / \tau\right)},
\end{equation}
where $\mathbf{h}_i$ are point features obtained by attaching a decoder to the encoder $f_\theta$, $\tau$ is a temperature parameter, and $\mathcal{P}$ is the set of matched points.

More pre-training details are provided in the supplementary materials, where we also review the same contrastive scheme applied at shape-level, rather than the point-level.

\section{Transferring to Downstream Tasks}
\label{sec:Evaluation}

\subsection{Approach}

In order to evaluate the effectiveness of supervised pre-training relative to self-supervised methods, such as contrastive learning, we employ both \emph{linear probing} and \emph{fine-tuning} strategies of pre-trained models across various architectures, datasets, and tasks. Linear probing assesses the quality of the representations learned during pre-training by training a linear classifier on top of the frozen backbone weights, while fine-tuning involves adjusting all weights of pre-trained models to new tasks or datasets. Our first goal is to establish a comprehensive comparison for supervised vs. contrastive pre-training, which has been largely omitted in existing literature \cite{wang2021occo,sauder2019self,afham2022crosspoint}, and to identify the scenarios in which supervised pre-training can be beneficial. 

In the full fine-tuning setting, we adhere to standard learning practices for each architecture. The precise details of the fine-tuning parameters, including learning rates, batch sizes, and optimization algorithms, are documented in the supplementary materials.

\begin{figure}[t!]
  \centering
  \begin{subfigure}[b]{0.48\linewidth}
    \includegraphics[width=\linewidth]{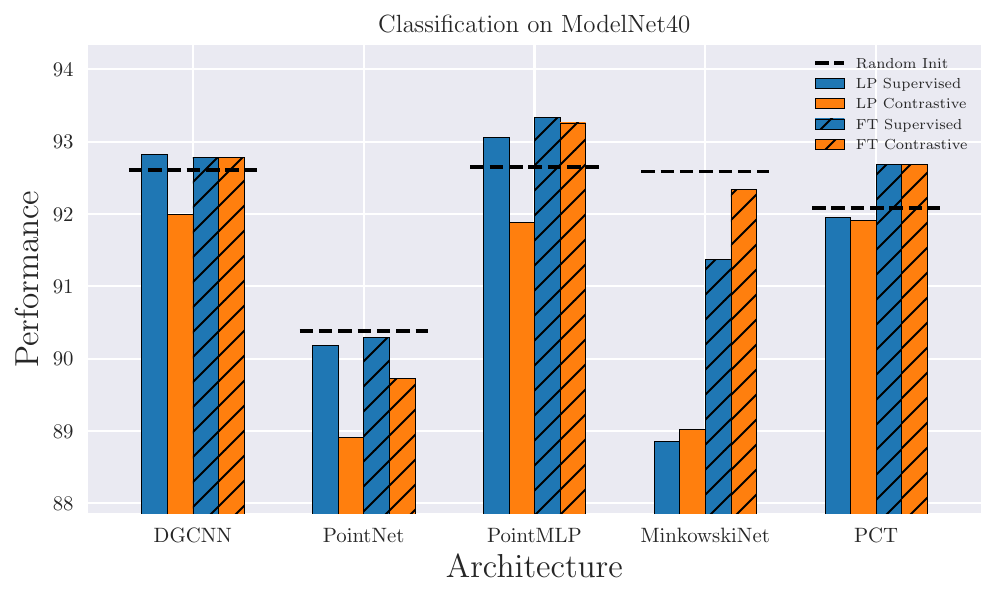}
    \caption{ModelNet40}
    \label{fig:eval_results_m40}
  \end{subfigure}
  \hfill
  \begin{subfigure}[b]{0.48\linewidth}
    \includegraphics[width=\linewidth]{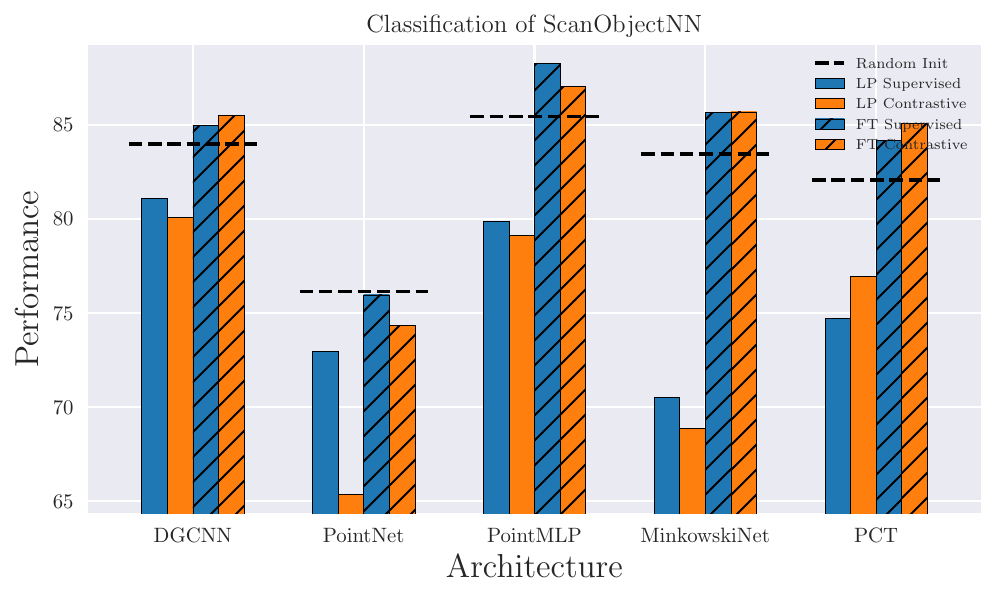}
    \caption{ScanObjectNN}
    \label{fig:eval_results_sonn}
  \end{subfigure}
  \caption{Evaluation on (a) ModelNet40 and (b) ScanObjectNN classification tasks of different pre-trained models, using linear probing (LP -- solid bars) and fine-tuning (FT -- dashed bars) settings. Random Init is a randomly initialized model. Transfer learning performance depends on pre-training scheme, architecture and evaluation protocol.}
  \label{fig:eval_combined}
\end{figure}

\subsection{Results}
\label{sec:results}

Our first set of results are summarized in Figures~\ref{fig:eval_results_m40},~\ref{fig:eval_results_sonn} and Table~\ref{tab:s3dis_table}. Below, we discuss several key observations. 

\mypara{Transfer learning via linear probing.} We first note that supervised pre-training consistently demonstrates superior performance across the majority of architectures in the linear probing setting (solid bars in Figures~\ref{fig:eval_results_m40},~\ref{fig:eval_results_sonn}). This suggests that supervised pre-training \textit{can} produce general-purpose, discriminative features that are useful even under fairly significant domain shift (e.g., synthetic shapes in ShapeNet vs. real scans in ScanObjectNN). Interestingly, linear probing with supervised pre-training can even surpass results obtained with full fine-tuning when the domain shift is small, as seen in ModelNet40 (Figure~\ref{fig:eval_results_m40}). We find the utility of supervised pre-training in this scenario noteworthy as it can refine the common belief that contrastive learning leads to more ``universally useful'' features.

\itpara{Unique architectural properties.} While supervised pre-training generally leads to useful final layer features, the results significantly depend on the choice of architectures. Specifically, we observe that the hierarchical nature of DGCNN and PointMLP results in more general features compared to global architectures such as PointNet. In addition, PCT's transformer architecture is adept at capturing global dependencies which overfits less with unsupervised (contrastive) pre-training.

\begin{table}[tb]

  \centering
      \caption{Evaluation on S3DIS semantic segmentation fine-tuning of pre-trained models. IoU metric used. \textit{Random init} is a randomly initialized model.
  }
    \label{tab:s3dis_table}
  \begin{tabular}{@{}lccc@{}}
    \toprule
    Architecture & Random Init & Supervised &  Contrastive\\
    \midrule
    DGCNN & 49.82 & 49.07 & \textbf{49.99} \\
    PointNet &\textbf{ 46.21} & 37.46 & 43.48 \\
    PointMLP & 56.59 & 55.7 & \textbf{58.00} \\
    MinkowskiNet & \textbf{66.89} & 64.07 & 60.97 \\
    PCT & 50.8 & 50.7 & \textbf{52.34} \\
  \bottomrule
  \end{tabular}

\end{table}

\mypara{Transfer learning with fine-tuning.}
We observe a notable shift of behavior when performing full fine-tuning, compared to linear probing, as shown in Figures~\ref{fig:eval_results_m40},~\ref{fig:eval_results_sonn} (dashed bars). We remark that for several architectures, such as DGCNN, MinkowskiNet and PCT, contrastive pre-training leads to better results compared to supervised learning under full fine-tuning (with a $\sim$0.1\% accuracy advantage in the on-par scenarios). This suggests that the utility of pre-trained models highly depends on \textit{how} those models are used in the downstream tasks. Specifically, we note that the utility of final layer features (such as with linear probing) is not always a good predictor for the performance under fine-tuning, as shown, e.g., in the case of DGCNN. In Section~\ref{sec:AnalysisGradientNorm} we analyze this behavior in depth and observe the critical role of \textit{early layers} and their ability to adapt to downstream data. Our analysis also highlights that the types of solutions obtained with supervised and contrastive pre-training are different, making a distinction between generalizable (features ready to be used in multi-task settings) and adaptable features.

In addition, we note that for most architectures, including MinkowskiNet, supervised pre-training is \textit{not} an ``upper bound'' to information gain from pre-training, differently from the claim made in~\cite{xie2020pointcontrast}, and that other pre-training strategies can lead to better results.

\itpara{Unique architectural properties.} PointNet, with its global feature aggregation strategy, tends to overfit to the source training dataset, and is thus not able to extract information that is useful for fine-tuning. MinkowskiNet's performance is influenced by variations in point cloud density and scale. Its ability to process large-scale scenes explains its high semantic segmentation performance (Table~\ref{tab:s3dis_table}), but highlights its dependence on the sampling density for effective transfer learning.

\section{Feature Analysis}
\label{sec:AnalysisFeatureActivations}

The inconsistency in the effectiveness of pre-training approaches across transfer learning scenarios leads us to explore the intrinsic properties that contribute to the transferability of model features. Our focus is particularly on the behavior of early layers, given their established importance in 2D transfer learning \cite{yosinski2015understanding}.

\begin{figure}[tb]
  \centering
  \begin{subfigure}{0.46\linewidth}
    \includegraphics[width=\linewidth]{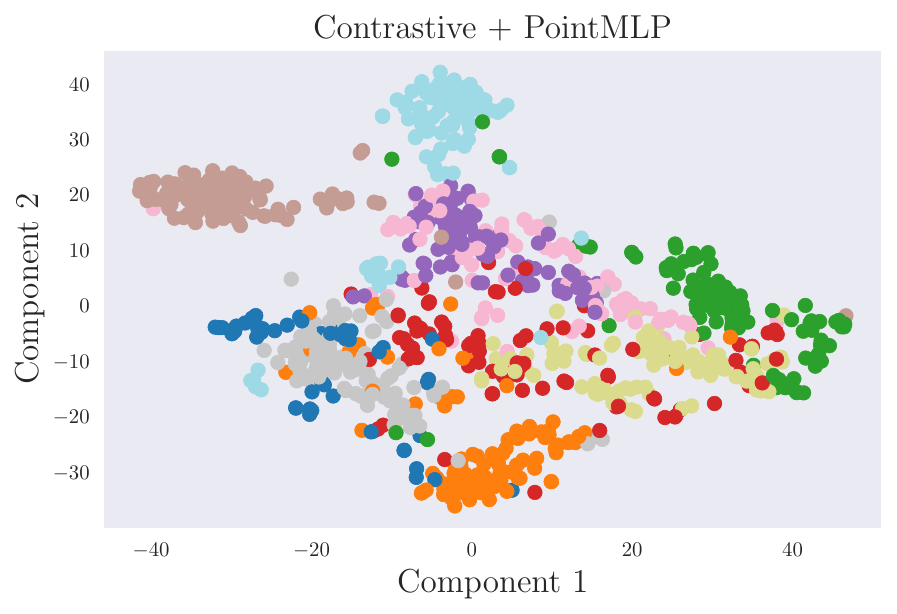}
  \end{subfigure}
  \hfill
  \begin{subfigure}{0.46\linewidth}
    \includegraphics[width=\linewidth]{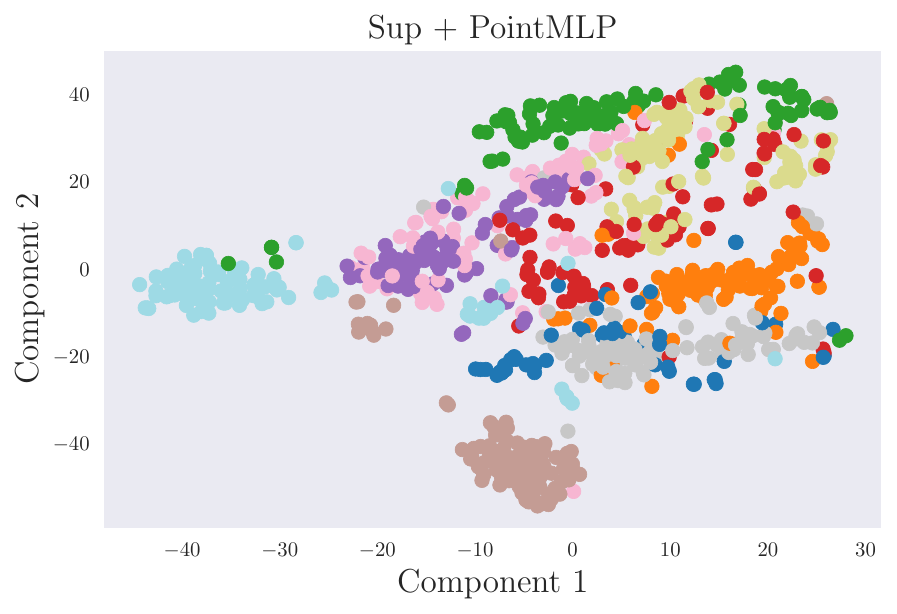}
  \end{subfigure}
  
  \begin{subfigure}{0.46\linewidth}
    \includegraphics[width=\linewidth]{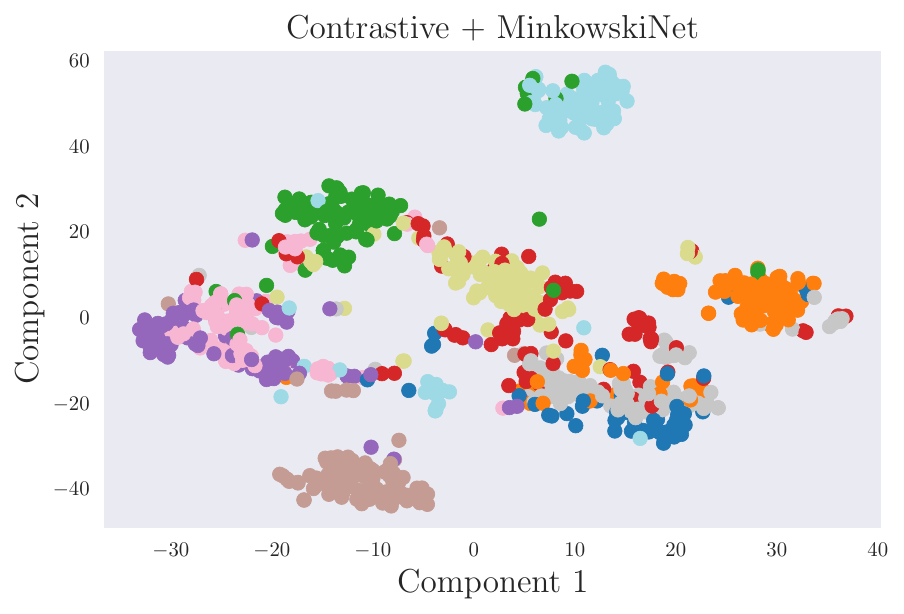}
  \end{subfigure}
  \hfill
  \begin{subfigure}{0.46\linewidth}
    \includegraphics[width=\linewidth]{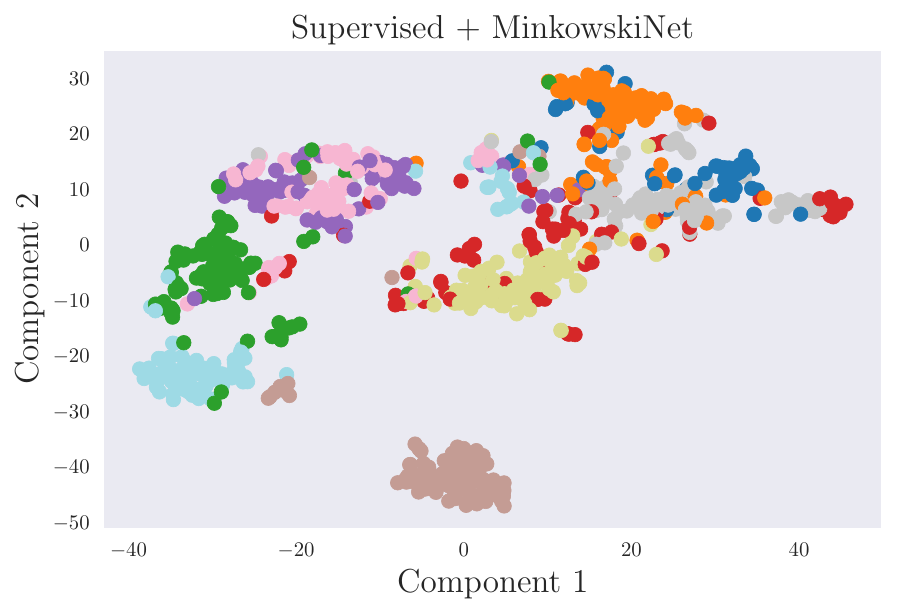}
  \end{subfigure}
  \caption{t-SNE plots of the first-layer feature activation for different architectures and pre-training schemes. We use the \textit{ModelNet10} evaluation set, which is a subset of ModelNet40 containing 10 classes, each represented by a different color. \textit{Clusters} are formed even in the feature space of first layers, which implies their discriminative capability. Visualization on additional architectures can be found in the supplementary.}
  \label{fig:tsne}
\end{figure}

\begin{table}[tb]
  \label{tab:layer_evaluation_results}
  \centering
  \caption{Linear SVM classification of features extracted from the first and last layers for different architectures and pre-training strategies. Downstream task is classification on ModelNet40. The 1st layer's evaluation displays remarkably high \textit{accuracy}, not that far from the last layer's evaluation accuracy.}
  \begin{tabular}{@{}lcccc@{}}
    \toprule
    Architecture & \multicolumn{2}{c}{Supervised } & \multicolumn{2}{c}{Contrastive} \\
    \cmidrule(lr){2-3} \cmidrule(lr){4-5}
     & 1st Layer & Last Layer & 1st Layer & Last Layer \\
    \midrule
    DGCNN & 81.80 & 90.92 & 81.15 & 89.87 \\
    PointNet & 79.37 & 88.24 & 79.29 & 86.46 \\
    PointMLP & 82.69 & 91.32 & 74.59 & 89.7 \\
    MinkowskiNet & 77.51 & 88.4 & 70.74 & 87.72 \\
    PCT & 78.9 & 90.55 & 77.9 & 89.87 \\
  \bottomrule
  \end{tabular} 
  \label{tab:linear_svm}
\end{table}

\mypara{Discriminative capability of early layers.} 
To illustrate the behavior of the early layers, we plot a t-SNE visualization of the first-layer features of pre-trained models in Figure~\ref{fig:tsne} on the ModelNet10 dataset. We observe the appearance of discernible clusters even in these early layers and without any fine-tuning. This clustering in the feature space suggests that \textit{early layers} possess a class-discriminative capacity. This points to the potential of early layers to contribute significantly to the downstream task, unlike in the 2D domain where early layers are typically "universal" and do not need to be adapted \cite{matsoukas2022makes}. 

To analyze this behavior quantitatively, we used a linear SVM to evaluate the utility of early layer features against those from output layers. The results, presented in Table~\ref{tab:linear_svm} highlight the remarkable classification performance of the early layers even with a simple linear probing on a new downstream dataset. This behavior is observed for all studied architectures. As a point of comparison, we evaluated the \textit{first layer} of the EfficientNetB0 \cite{tan2019efficientnet} backbone pre-trained with ImageNet \cite{deng2009imagenet} on the Oxford-IIIT Pet Dataset \cite{oxford}, resulting in 10\% classification accuracy (against  92.28\% obtained using the output layers), which showcases the low discriminative capability of early layers in the 2D domain.

As observed in Section \ref{sec:Evaluation}, the utility of final layer features is not always a good predictor for the performance under fine-tuning for 3D transfer learning. Coupled with the discriminative power of early layers revealed above, this leads us to investigate how the network weights change and what factors contribute to successful fine-tuning. 

\begin{figure}[tb]
  \centering
  \begin{subfigure}{0.48\linewidth}
    \includegraphics[width=\linewidth]{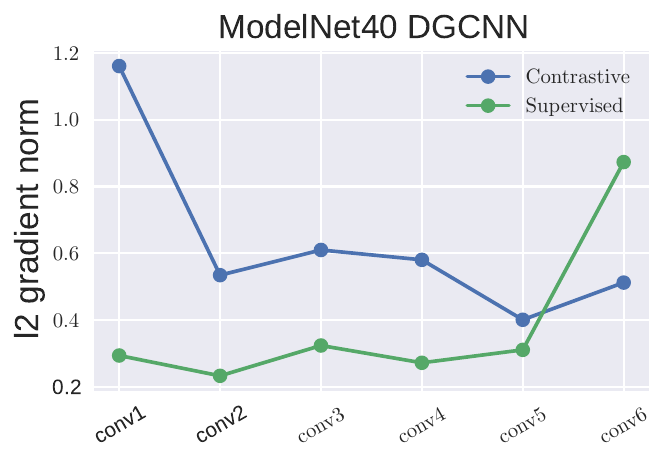}
  \end{subfigure}
  \begin{subfigure}{0.48\linewidth}
    \includegraphics[width=\linewidth]{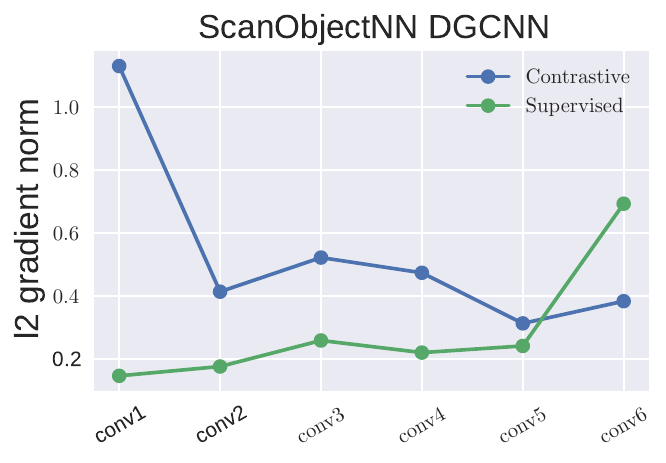}
  \end{subfigure}
  \\
  \begin{subfigure}{0.48\linewidth}
    \includegraphics[width=\linewidth]{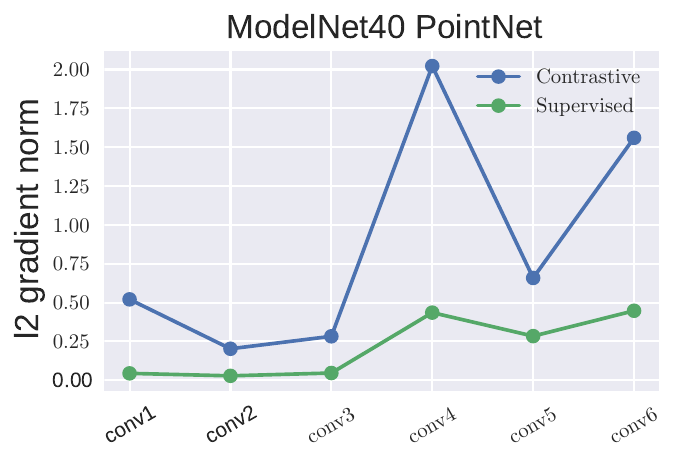}
  \end{subfigure}
  \begin{subfigure}{0.48\linewidth}
    \includegraphics[width=\linewidth]{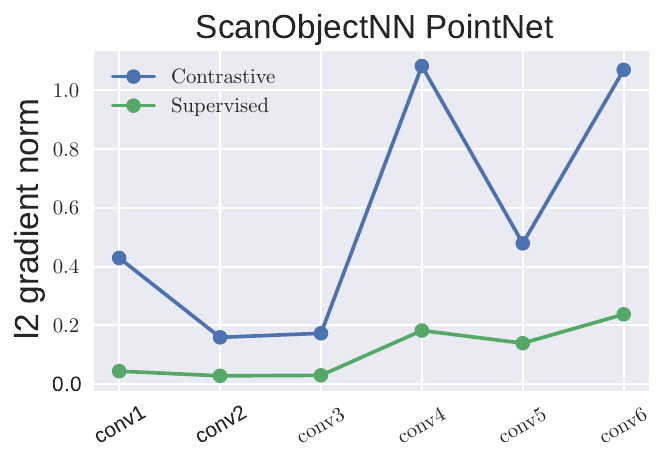}
  \end{subfigure}
  \caption{Layer-wise gradient norms of pre-trained models on downstream ModelNet40 and ScanObjectNN datasets using DGCNN and PointNet. Supervised pre-training shows low gradient norms, especially in early layers. Further analysis across architectures is provided in the supplementary.}
  \label{fig:grads}
\end{figure}

\mypara{Gradient norm of layer weights.} 
\label{sec:AnalysisGradientNorm}
To understand how the models behave in the fine-tuning setting, we analyze the norm of the gradients of pre-trained networks with respect to downstream data. Figure \ref{fig:grads} shows the layer-wise gradient norm, computed using a cross-entropy loss on the downstream training data using frozen pre-trained weights. We observed that for contrastive pre-training, early layers exhibit a higher gradient norm compared to the last layer. This finding further substantiates the need, not only for \textit{last layers}, but also for \textit{early layers} to adapt to new data/tasks when dealing with 3D data for successful transfer learning.

As shown in Figure~\ref{fig:grads} supervised pre-trained models demonstrate a low gradient norm, with the disparity between supervised and contrastive pre-training being more prominent especially in the early layers. The low gradient norm in supervised pre-training suggests that the learned features are confined to a local region of the optimization landscape. This implies that these features are \textit{less prone to adaptation when exposed to new data or tasks} during fine-tuning, which aligns with our previous observations regarding the limited \textit{adaptability} of supervised features under significant data shifts (Section~\ref{sec:results}). In contrast, features developed through contrastive pre-training, particularly in the early layers, show greater potential for adapting to new inputs, thereby enhancing their success in fine-tuning.

We note that this adaptability is not the same as feature generality, since the latter typically applies only to the final layers, which, as shown in Section~\ref{sec:Evaluation} can be informative (in linear probing) despite not being adaptable (in fine-tuning), e.g., for supervised pre-training.

\section{Layer-wise Geometric Regularization}
\label{sec:regul}

\subsection{Normal Prediction}
The qualitative insights of our analysis have revealed various challenges in pre-training that can hinder effective transfer learning. Specifically, we have observed a correlation between low gradient norms in early layer weights and diminished fine-tuning performance, indicating a decreased adaptability of the learned features when moving from source to downstream tasks.

Given these insights, our objective is to improve the standard pre-training method while adhering to the pre-text task. We achieve this by implementing a straightforward regularization strategy designed to enhance the geometric and, consequently, more universally applicable characteristics of the learned features.

As highlighted in our analysis, effective regularization to pre-training should promote low-level features, particularly in early layers, that can adapt better to unseen data.
Besides, the regularization needs to leverage properties intrinsic to input point clouds. We find that predicting geometric properties as an auxiliary task aligns with these criteria, as it promotes \textit{local,} category-agnostic attributes of input point clouds. A practical method to achieve this is to predict point-wise normal vectors, a task that can be achieved without extra labels by using estimation methods such as PCA, or by extracting them from raw meshes when available.

We apply this regularization only to a set of early layers, whereas the output features of the entire architecture continue to serve the primary pre-text task. This strategy infuses purely geometric information into early layers with the goal of making them universally applicable and reducing the dependency to the pre-text task. 
Let $H_{\text{normal}}(\mathbf{f}_{l})$ denote the normal prediction head, $\mathbf{f}_{l}$ the point-wise features of layer $l$, and $\mathbf{n}$ the ground truth normals. The combined training objective comprises the pre-training loss $L_{\text{pre-train}}$ and the regularization loss $L_{\text{regul}}$. We use absolute cosine similarity in $L_{\text{regul}}$ since estimated normals are not oriented. Our total pre-training loss can be formulated as:
\begin{equation}
    L_{\text{total}} = L_{\text{pretrain}} + L_{\text{regul}}\left(H_{\text{normal}}(\mathbf{f}_{l}), \mathbf{n}\right).
\end{equation}

\subsection{Evaluation}

\begin{table}[tb]
  \caption{Evaluation of geometric regularization of supervised and contrastive pre-training. Evaluations on Shape classification on ModelNet40 and ScanObjectNN, and 3D scene segmentation on S3DIS with DGCNN architecture. Regularization improves downstream performance of supervised pre-training in the fine-tuning setting.
  }
  \label{tab:regul}
  \centering
  \begin{tabular}{@{}lccc@{}}
    \toprule
    Pre-training & ModelNet40 & ScanObjectNN & S3DIS \\
    \midrule
    Supervised & 92.78 & 84.94 & 49.07 \\
    Supervised + regularization & \textbf{93.34} & \textbf{85.91} & \textbf{49.88} \\
    \midrule
    Contrastive  & 92.78 & 85.05 &  49.99 \\
    Contrastive + regularization  & \textbf{93.18} & \textbf{85.39} &  \textbf{50.44} \\
  \bottomrule
  \end{tabular}
\end{table}

\begin{figure}[tb]
  \centering
  \begin{subfigure}{0.45\textwidth}
    \includegraphics[width=1\linewidth]{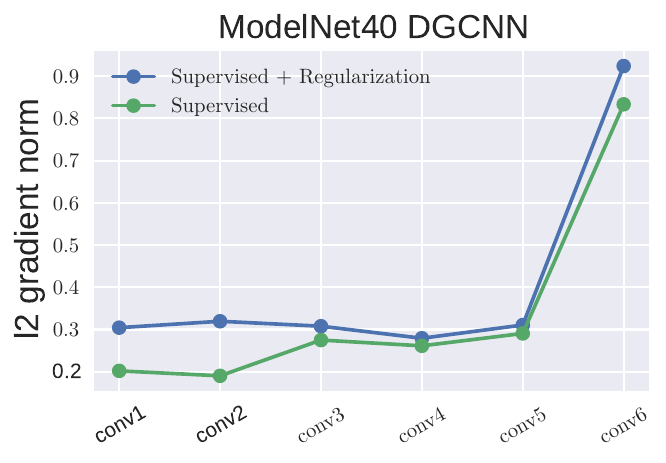}
  \end{subfigure}
  \begin{subfigure}{0.45\textwidth}
    \includegraphics[width=1\linewidth]{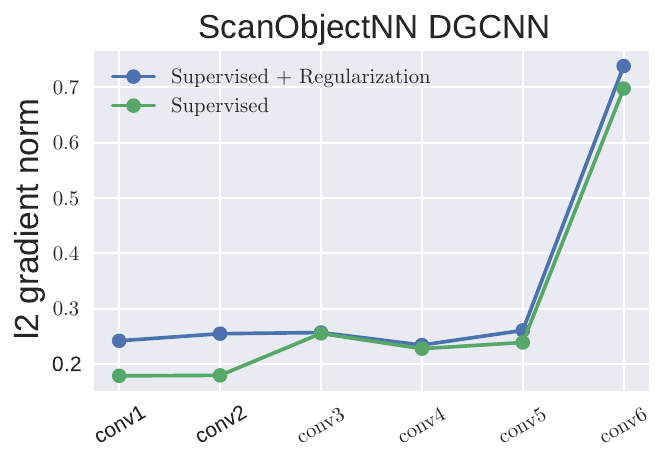}
  \end{subfigure}
  \caption{Layer-wise gradient norms of supervised pre-trained models with and without regularization on downstream ModelNet40 and ScanObjectNN datasets using DGCNN. Regularization increases the gradient norm values, especially in early layers.}
  \label{fig:grad_regul}
\end{figure}

The experimental setup for pre-training and fine-tuning remains consistent with our previous evaluation. We use normals computed through PCA on local neighborhoods comprising 30 points within point clouds of size 2048. We explore the optimal number of layers required for the most effective regularization, and find that regularizing the features of the first two layers yields the best results. Through empirical testing to determine the optimal layers for regularization, we discovered that targeting early layers for regularization yields better outcomes than applying it to the encoder’s output layer alone. Ablation of layer choice and other architectures is included in the supplementary.

When comparing between baseline and regularized supervised pre-training, Table \ref{tab:regul} shows notable improvement (1\% gain) in all evaluated downstream data and tasks. This improvement is also correlated to an increase in gradient norm values as shown in Figure~\ref{fig:grad_regul}. These  results highlight the potential of layer-wise regularization and encourages its use for future supervised pre-training.

Despite the primary goal of addressing the limitations of early layers in supervised pre-training through this regularization technique, we also explore its application in other pre-text tasks, such as contrastive learning. According to Table \ref{tab:regul}, incorporating our regularization into contrastively pre-trained models can benefit fine-tuning especially under strong domain shift. This suggests that layer-wise regularization can be a useful general approach for diverse transfer learning strategies.

\section{Summary and Key Takeaway Remarks}

Our findings emphasize several critical aspects of pre-training strategies and their impact on transfer learning performance across different architectures.

The choice of evaluation protocol is critical. Supervised pre-training excels in linear probing often surpassing contrastive pre-training. This highlights its potential for universal task ready features. However, this approach falls behind contrastive learning in fine-tuning, emphasizing the importance of feature \textit{adaptability}.

The architecture also influences pre-training outcomes. We find that simpler models that capture global features such as PointNet fail at transferring learned knowledge when fine-tuned. More local point-based methods such as PointMLP benefit greatly from pre-training, especially with supervised pre-training. Transformers benefit more from contrastive learning. Complex architectures that process large scale data such as MinkowskiNet, which is based on sparse convolutions can boost the transfer learning performance under small domain shift (\eg{} scenes to scenes, shapes to shapes) but can be sensitive to changes in data sampling. Graph-based such as DGCNN with local and global feature extraction, can be pre-trained and fine-tuned efficiently, although they do not always exhibit the best overall performance.

Second, we observe that unlike the 2D case, features learned \textit{even in the earliest layers} of 3D architectures tend to be class-specific, which can hinder their utility in transfer learning. This highlights the importance of adaptability of early layers for point cloud transfer learning.

Finally, regularizing exclusively early layers using geometric signals can improve pre-training in several settings, particularly for supervised pre-training. This approach addresses the lack of general-purpose low-level 3D features. This insight also opens the door for future research to focus on early layer-wise regularization rather than applying it uniformly across the entire model.

\section{Conclusion, Limitations and Future Work}
\label{sec:conclusion}

In this paper, we have conducted the first in-depth qualitative and quantitative investigation of the key factors when performing point cloud transfer learning with supervised vs contrastive pre-training and proposed a regularization approach informed by this analysis. We evaluated several architectures under pre-training approaches and found that their performance changes with the transfer learning protocol. We also examined early layer importance through their discriminative capability and fine-tuning adaptability, shedding light on their utility and importance for point cloud transfer learning. Finally, we correlated this analysis with a new regularization approach that targets early layers to improve on downstream performance. Overall, our work establishes a consistent evaluation framework, presents detailed analysis tools, and proposes an appropriate simple method for successful transfer learning on point cloud data, which can both inform and allow to compare future designs. 

As our main focus was on the careful analysis of the factors that contribute to successful transfer learning for point clouds, we did not investigate \textit{all} possible combinations of the source datasets, architectures, pre-training strategies, fine-tuning approaches, and downstream tasks. The resulting combinatorial complexity would incur very significant computational costs, and furthermore might obscure the possible analytical insights. Nevertheless, in the future, an analysis of other tasks and pre-training strategies can fit within and complement the framework that we established. Finally, as suggested by our analysis, there is currently a need for new architectures and pre-training approaches, that are robust under changes of sampling density and can lead to multi-scale features, which would generalize and adapt to new downstream data. Exploring the possible solution space is an exciting direction for future work.

\mypara{Acknowledgements} Parts of this work were supported by the ERC Starting Grant 758800 (EXPROTEA), ERC Consolidator Grant 101087347 (VEGA), ANR AI Chair AIGRETTE, as well as gifts from Ansys and Adobe Research.

%
%
\bibliographystyle{splncs04}
\bibliography{main}

\newpage

This supplementary document provides additional details and results about the frameworks employed in our work. Specifically, we first describe the architectures used in our experiments in \cref{supp_sec:Architectures}. Then we detail in \cref{supp_sec:pretraining} other pre-training approaches evaluated in our study with their respective configurations. In \cref{supp_sec:evaluation}, we describe our transfer learning protocols and the corresponding optimization parameters. In \cref{supp_sec:analysis}, we extend the analysis of the discriminative ability of early layers and gradient norm to other configurations. Finally, in \cref{supp_sec:regul} we provide additional results on our geometric regularization approach. 
\section{Architectures}
\label{supp_sec:Architectures}

The architectures used in our experiments were selected to be as similar as possible to original implementations. Some minor changes were made in order to properly implement pre-training strategies and to keep a consistent encoder architecture across them. These modifications were not intended to tailor the architectures to get the best performance, but just to keep consistency.

Unless specified otherwise, encoders share a feature embedding dimension of 1024 which is projected to a 128 per-point feature dimension for point-level contrastive-learning and 256 feature dimension for shape-level contrastive learning (see \cref{supp_sec:pretraining}).

\textbf{DGCNN:} For both pre-training and fine-tuning, we employed the six-layer deep part-segmentation architecture as proposed in the original DGCNN work \cite{wang2019dynamic}.

\textbf{PointNet:} Modifications were applied to the PointNet architecture to achieve a uniform encoder suitable for both classification and segmentation tasks, resulting in slight performance differences from the original implementation.

\textbf{PointMLP:} We used the standard (non-elite) version of the original PointMLP \cite{ma2022rethinking} implementation of the encoder. Although no benchmark or implementation was initially provided for the semantic segmentation task, we add a simple decoder using the PointNet feature propagation module implemented in PointNet++ \cite{qi2017pointnet++} to evaluate PointMLP on the semantic segmentation downstream task, enable point-level contrastive learning and apply layer-wise geometric regularization.  

\textbf{MinkowskiNet:} We adopt the SR-UNet architecture originally proposed in \cite{choy20194d} and used in PointContrast \cite{xie2020pointcontrast}. Though this architecture is originally tuned for segmentation tasks, its encoder can be effectively used for classification tasks. We take 32 as the point feature embedding dimension, to be consistent with PointContrast implementation.

\textbf{PCT:} We adopt the baseline PCT architecture \cite{guo2021pct} used for classification (with downsampling module) for the classification downstream tasks. For the segmentation task, we implement a version without downsampling as proposed in the original paper since the implementation was missing. Each variation underwent its own pre-training process.

Maintaining architectural consistency while exploring various pre-training approaches posed significant challenges. Consequently, we prioritized maintaining consistent parameterization across architectures over achieving the highest possible performance.

\section{Pre-training}
\label{supp_sec:pretraining}

In addition to the supervised and point-level contrastive approach presented in the main paper, we provide additional results on two more pre-training strategies: 

\begin{itemize}
    \item \textbf{Shape-level contrastive learning (SC):} Inspired by the commonly used contrastive learning approach in 2D \cite{chen2020simple}, which contrasts between entire images, we apply the same view generation technique used in the point-level contrastive learning approach (denoted as PC) to generate a positive pair of shapes. We add an MLP projection head to the encoder for contrastive loss computation.
    \item \textbf{Point-DAE:} We follow the Point-DAE work \cite{zhang2022point}, which investigates denoising auto-encoders for self-supervised pre-training on 3D point clouds. They propose several corruption settings, but we select the affine + masking corruption for its performance and generalizability and evaluate if observations made in our work are still relevant for this type of pre-training. One advantage of this simple scheme is that it can be used with different type of 3D backbones, unlike other masked pre-training methods that are exclusive to vanilla or vision transformers (\eg Point-MAE \cite{pang2022masked})
\end{itemize}

Efforts were made to standardize pre-training settings across architectures and methods to reduce biases due to parameter differences.

Although we focus on contrastive vs. supervised pre-training, our evaluations span supervised, point-level, and shape-level contrastive pre-training methods, alongside a reconstruction-based approach. We focus on single modality (3D) pre-training, but multi-modal pre-training methods can also be explored in future work for comparison.

When using contrastive pre-training, we introduce a set of invariances by applying data augmentation. We start by normalizing the point cloud and perform random geometric transformations, including translation with a magnitude of 0.5, scaling between 80\% and 125\%, and rotation of magnitude 45$^{\circ}$. These transformations and their values are typically used in data augmentation schemes for object-level datasets. We additionally simulate partial data to obtain complex shapes that are more challenging for the contrastive pretext task. This is done by cropping the original shape to a certain percentage. We experimented with several crop ratios ranging from $0.2$ to $0.8$ and found that $0.5$ gives the best evaluation performance. 

The ADAM optimizer \cite{kingma2014adam}, with a learning rate of $10^{-3}$ and a weight decay of $10^{-6}$, is used for 100 epochs, and a batch size of 32 is adopted. Although the number of epochs might be low for Point-DAE pre-training, it was chosen to maintain consistency in comparisons. Temperatures for SC and PC pre-training are set to 0.1 and 0.4, respectively, following conventions established in PointContrast \cite{xie2020pointcontrast}. For MinkowskiNet, a voxel size of 0.1 was selected after testing ranges from 0.1 to 0.5, showing optimal pre-training outcomes. Pre-training is conducted on the entire ShapeNetCoreV2 dataset \cite{shapenet2015}.

\section{Evaluation}
\label{supp_sec:evaluation}

\subsection{Settings}
We maintained consistent linear probing and fine-tuning optimization parameters across downstream classification tasks, utilizing an SGD optimizer with a learning rate of $10^{-3}$, a weight decay of $10^{-6}$, training for 200 epochs, and a batch size of 32. For semantic segmentation tasks on S3DIS, the batch size was adjusted to 24, and the training duration was set to 100 epochs. For MinkowskiNet, we applied a voxel size of 0.1 for shape classification and 0.5 for semantic segmentation on S3DIS.

Data augmentation for ModelNet40 classification included random geometric transformations such as translations (up to 0.2 units) and scalings (between two-thirds and one and a half times the original size), with additional rotations for ScanObjectNN classification. For MinkowskiNet, we adopted the PointContrast data augmentation scheme, with transformations excluded for semantic segmentation tasks.

In our study, linear probing was not conducted for semantic segmentation due to the absence of a pre-trained decoder for supervised and SC contrastive pre-training. This approach is less common for this task, as encoder weights typically generate feature embeddings for the entire input, not point-wise. For PC pre-training with DGCNN, keeping the encoder and decoder frozen while training only the last layer on S3DIS resulted in a significantly lower mIoU score (24.09\%) compared to fine-tuning (49.99\%).

\subsection{Additional results}

\begin{figure}[tb]
  \centering
  \begin{subfigure}[b]{0.48\linewidth}
    \includegraphics[width=\linewidth]{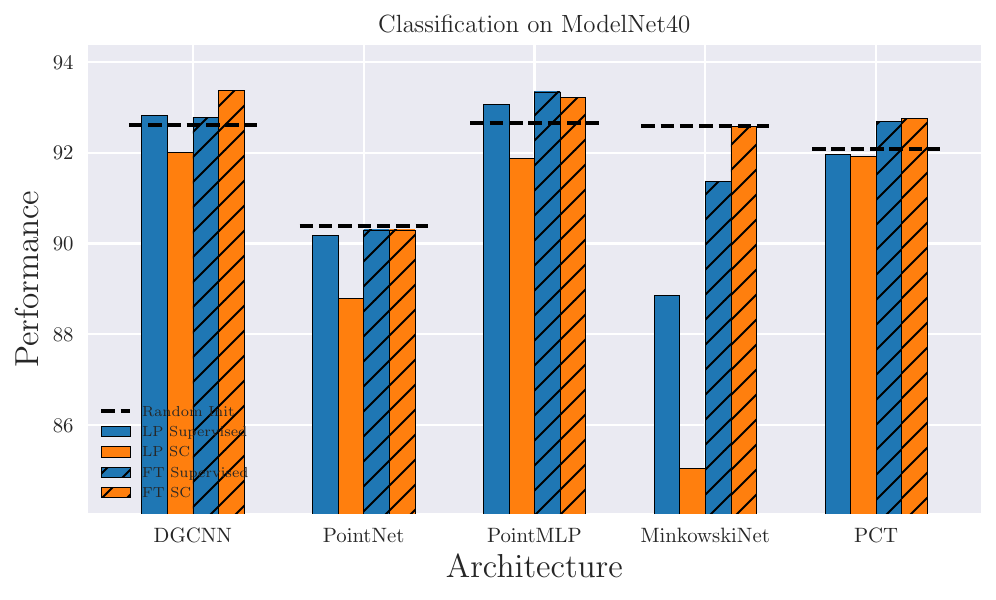}
    \caption{Modelnet40}
  \end{subfigure}
  \begin{subfigure}[b]{0.48\linewidth}
    \includegraphics[width=\linewidth]{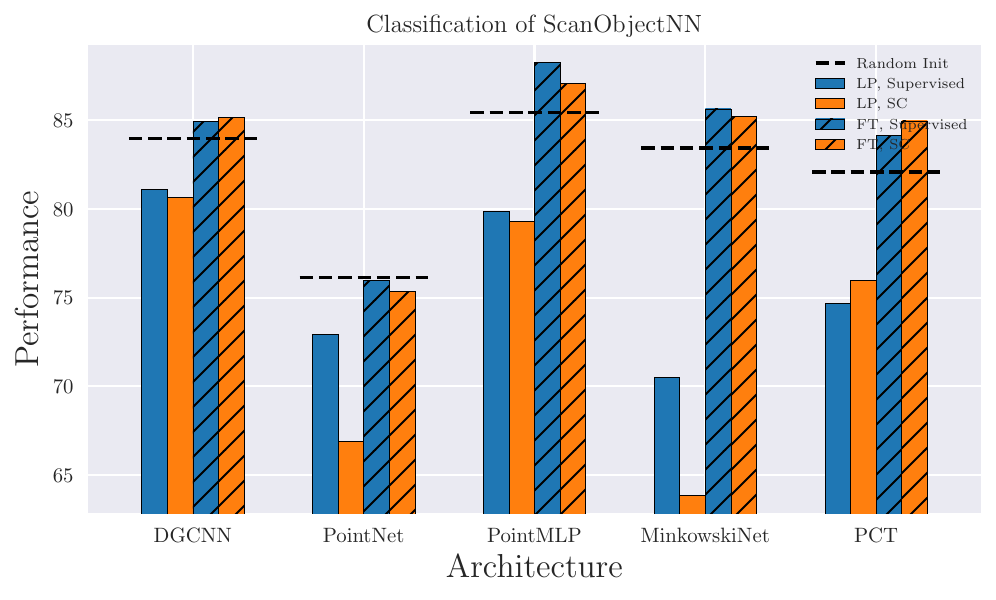}
    \caption{ScanObjectNN}
  \end{subfigure}
  \hfill
  \caption{Evaluation on (a) ModelNet40 and (b) ScanObjectNN classification tasks of SC pre-trained models in comparison to supervised pre-trained model, using linear probing (LP -- solid bars) and fine-tuning (FT -- dashed bars) settings. Random Init is a randomly initialized model.}
  \label{supp_fig:eval_sc}
\end{figure}

\begin{table}[ht!]
  \centering
  \caption{\textbf{Evaluation of denoising auto-encoder pre-training strategy.} Shape classification on ModelNet40 and ScanObjectNN, and 3D scene segmentation on S3DIS across different pre-training strategies and architectures. Accuracy metric used for classification and mIoU metric for semantic segmentation. Bolded results represent best evaluation metric for a specific dataset and architecture setting. RI is a randomly initialized model.}
  \begin{tabular}{lcccc}
    \toprule
    Pre-training Strategy & ModelNet40 \cite{wu20153d} & ScanObjectNN \cite{uy2019revisiting} & S3DIS \cite{armeni20163d} \\
    \midrule
    \multicolumn{4}{c}{Linear Probing} \\
    \midrule
    DAE + DGCNN & 92.00 &  72.31 &  -\\                
    \midrule
    DAE + PointMLP & 91.15 & 73.46 & -\\
    \midrule
    DAE + PCT & 89.33 &  65.82 & -\\
    \midrule
    \multicolumn{4}{c}{Fine-tuning} \\
    \midrule
    DAE + DGCNN & \textbf{92.69} & \textbf{84.98} &  \textbf{51.4} \\
    RI + DGCNN & 92.61 & 83.96 &  49.82 \\
    \midrule
    DAE + PointMLP & \textbf{93.18} & \textbf{85.5} & \textbf{59.24} \\
    RI + PointMLP & 92.65 & 85.43 & 56.59 \\
    \midrule
    DAE + PCT & \textbf{91.4} & \textbf{77.65} & \textbf{50.87} \\
    RI + PCT & 91.19 & 76.72 & 50.8\\
    \bottomrule
  \end{tabular}
  \label{supp_tab:evaluation-table}
\end{table}

\noindent\textbf{Shape-level contrastive pre-training (SC).} We evaluate transfer learning performance of shape-level contrastive learning in comparison to supervised pre-training in Figure \ref{supp_fig:eval_sc}. Shape-level contrastive learning follows a similar pattern to Point-level contrastive learning results in Figure~2 of the main paper. However, we observe that sparse architectures (i.e. MinkowskiNet) benefit less from SC compared to what was observed for PC.

\noindent\textbf{Reconstruction type pre-training.} Results in Table~\ref{supp_tab:evaluation-table} for Point-DAE, a denoising auto-encoder (with affine and masking corruptions), indicate modest linear probing performance but promising fine-tuning results, especially in semantic segmentation tasks, highlighting the potential of reconstruction-based pre-training. The superior performance in semantic segmentation is correlated to the nature of the pretext task, which is focused on the point-level rather than the shape-level, as for contrastive pre-training. We omit MinkowskiNet because of the variable number of points in an input making it incompatible with the baseline framework of Point-DAE, and PointNet because of its global architecture which overfits to the pretext task.

\begin{figure}[t]
  \centering
  \begin{subfigure}[b]{0.48\linewidth}
    \includegraphics[width=\linewidth]{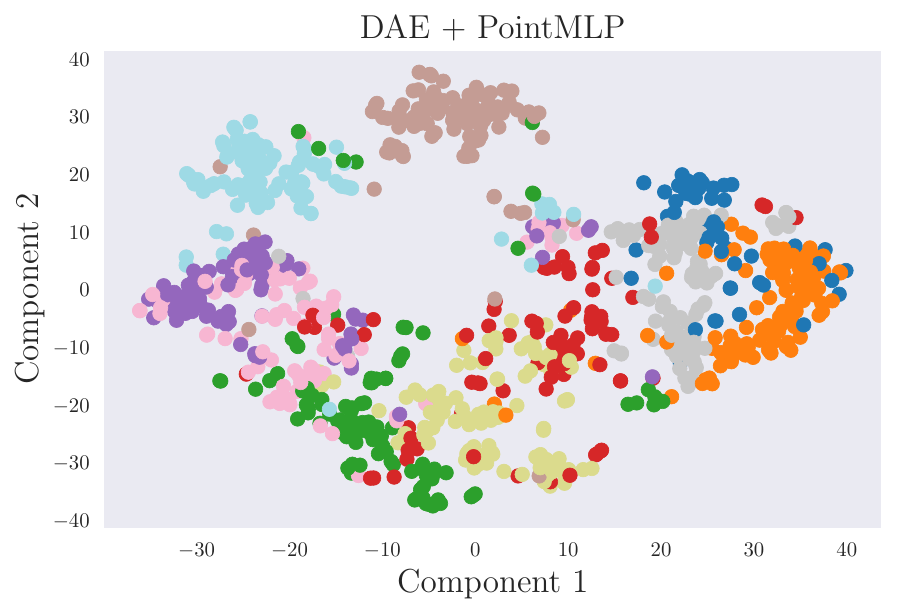}
    \caption{Point-DAE}
  \end{subfigure}
  \begin{subfigure}[b]{0.48\linewidth}
    \includegraphics[width=\linewidth]{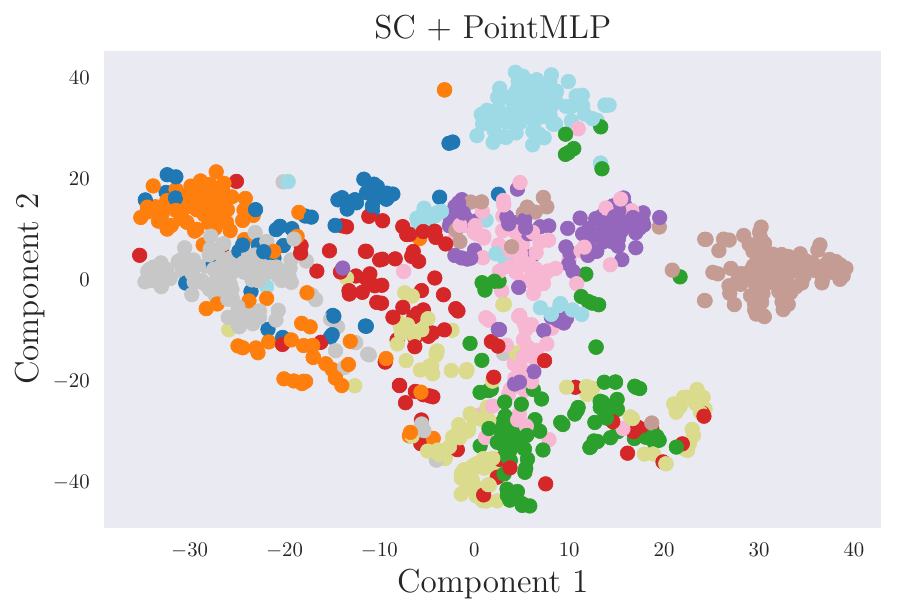}
    \caption{SC}
  \end{subfigure}
  \hfill
  \caption{\textbf{t-SNE plots of the first-layer feature activation for PointMLP under (a) Point-DAE and (b) SC pre-training.} Both pre-training strategies produce discriminative early layers.}
  \label{supp_fig:tsne_supp}
\end{figure}

\section{Analysis}
\label{supp_sec:analysis}

\textbf{Discriminative capability of early layers.} As depicted in Figure \ref{supp_fig:tsne_supp}, both shape-level contrastive learning (SC) and reconstruction-type pre-training (Point-DAE) lead to discernible clusters in feature space, indicating that discriminative early layers are common across pre-training strategies. This property is rarely observed in the 2D domain as discussed in the main paper.

\begin{figure}[ht]
  \centering
  \begin{subfigure}{0.32\textwidth}
    \includegraphics[width=1\linewidth]{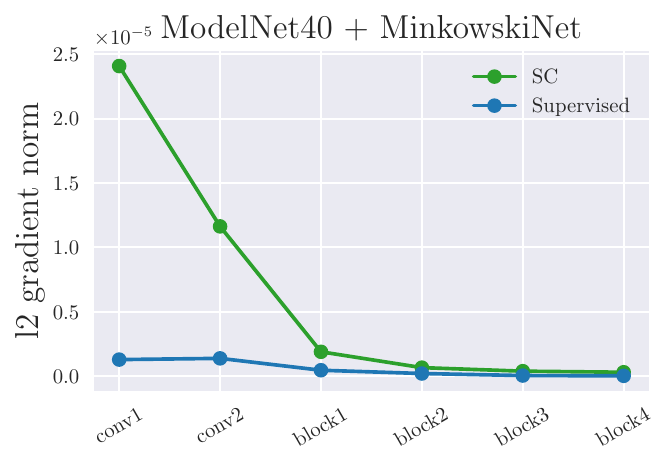}
  \end{subfigure}
  \begin{subfigure}{0.32\textwidth}
    \includegraphics[width=1\linewidth]{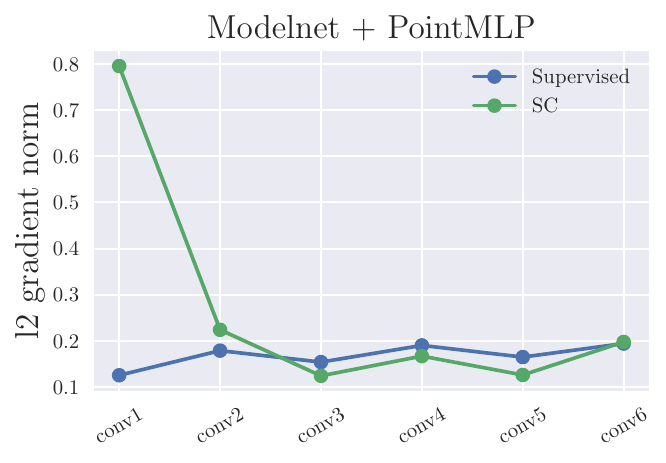}
  \end{subfigure}
  \begin{subfigure}{0.32\textwidth}
    \includegraphics[width=1\linewidth]{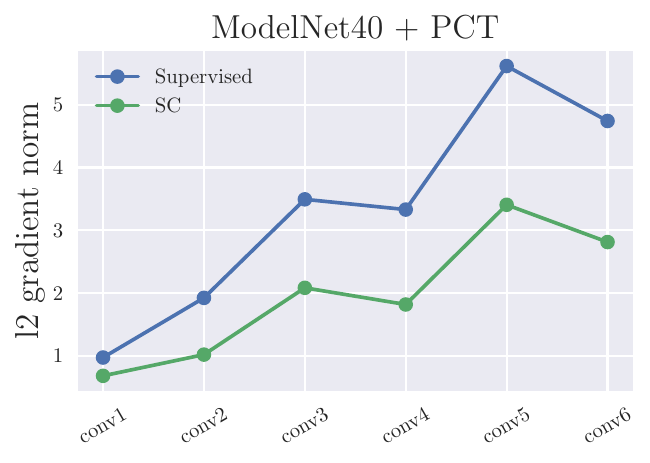}
  \end{subfigure}
  \\
  \begin{subfigure}{0.32\textwidth}
    \includegraphics[width=1\linewidth]{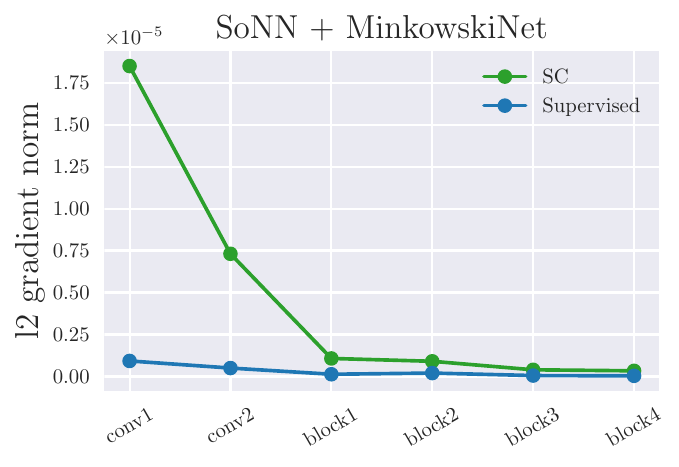}
  \end{subfigure}
  \begin{subfigure}{0.32\textwidth}
    \includegraphics[width=1\linewidth]{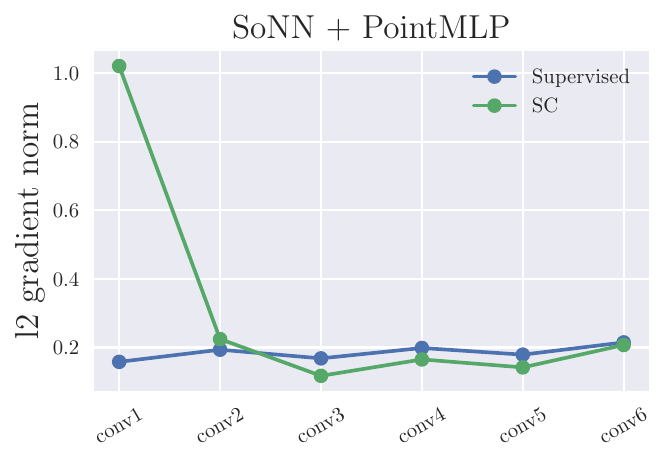}
  \end{subfigure}
  \begin{subfigure}{0.32\textwidth}
    \includegraphics[width=1\linewidth]{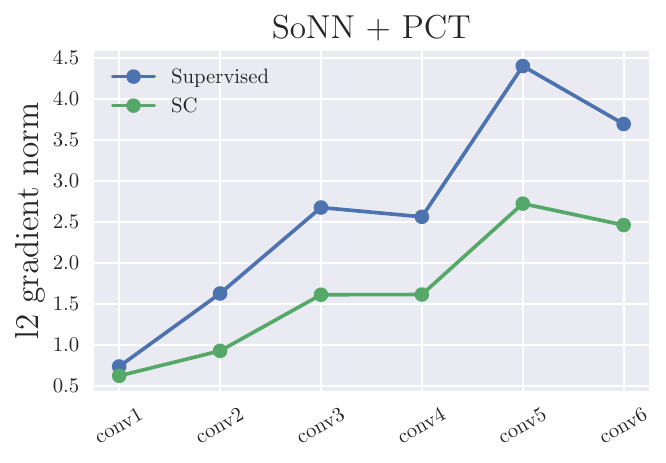}
  \end{subfigure}
  \caption{\textbf{Gradient norm analysis for pre-trained models.} Evaluation includes SC and Supervised pre-training across MinkowskiNet, PointMLP, and PCT. The $x$-axis represents convolutional layers at various depths.}
  \label{supp_fig:grad_sc}
\end{figure}

\begin{figure}[ht!]
  \centering
  \begin{subfigure}[b]{0.48\linewidth}
    \includegraphics[width=\linewidth]{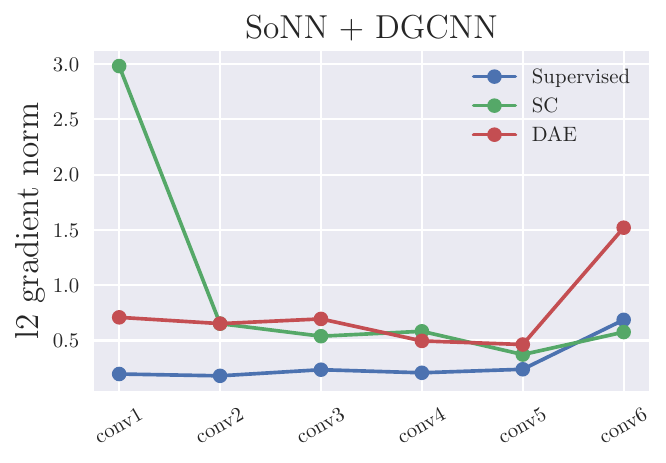}
  \end{subfigure}
  \hfill
  \caption{\textbf{Gradient norm analysis for pre-trained models.} Evaluation includes SC, Point-DAE and Supervised pre-training. The $x$-axis represents convolutional layers at various depths.}
  \label{supp_fig:grad_sonn_dae}
\end{figure}

\noindent\textbf{Gradient norm of layer weights for other architectures.} We generalize the results found in the main paper on gradient norm of layer weights for different architectures and other pre-training strategies. First, we see in Figure \ref{supp_fig:grad_sc} that supervised pre-training produces low gradient norm value for first layers across all studied architectures, although the exact pattern of the gradient norm curve changes through different convolutional layers from architecture to another. Second, these results correspond to shape-level contrastive learning, where the results resemble the point-level contrastive learning one. Third, reconstruction-type pre-training strategies like Point-DAE result in higher gradient norm values, as shown in Figure \ref{supp_fig:grad_sonn_dae}, but still lower than those from contrastive learning strategies, potentially explaining the reduced performance of Point-DAE.

\begin{figure}[ht!]
  \centering
  \begin{subfigure}{0.45\textwidth}
    \includegraphics[width=1\linewidth]{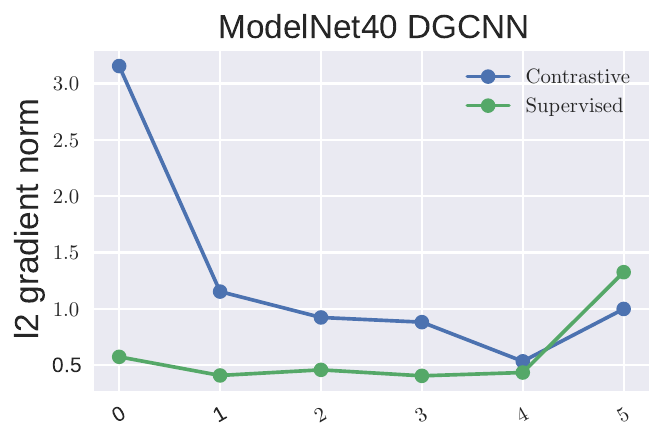}
  \end{subfigure}
  \begin{subfigure}{0.45\textwidth}
    \includegraphics[width=1\linewidth]{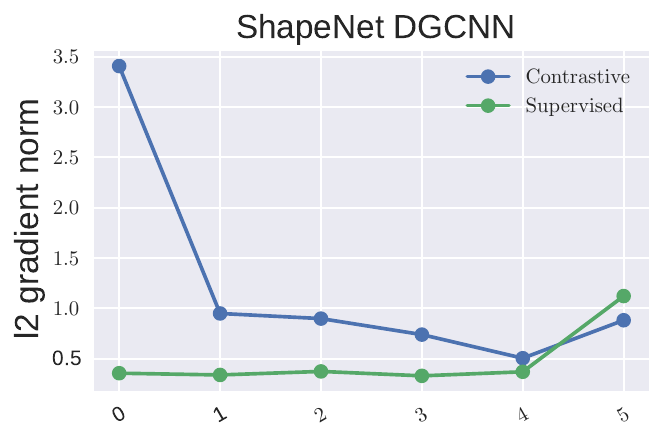}
  \end{subfigure}
  \caption{Layer-wise gradient norm of a model pre-trained on ScanObjectNN for two different downstream datasets.}
  \label{supp_fig:grads}
\end{figure}

\noindent\textbf{Gradient norm of layer weights for other source data.} We highlight that our key observations on the difference of adaptability between pre-training methods are not unique to the ShapeNet source data. As shown in Figure~\ref{supp_fig:grads}, in the setting where the pre-training data is ScanObjectNN, early layers are also less adaptable for supervised pre-training compared to contrastive pre-training.

\begin{table}[tb]
    \caption{\textbf{Fine-tuning of geometric regularized pre-trained models on different downstream data/tasks.} Shape classification on ModelNet40 and ScanObjectNN with more architectures. Regularization can improve downstream performance of supervised pre-training.}
  \centering
  \begin{tabular}{lcccc}
    \toprule
    Pre-training Strategy & PointNet & MinkowskiNet & PointMLP & PCT\\
    \midrule
    \multicolumn{4}{c}{ModelNet40 accuracy} \\
    \midrule
    Supervised  & 90.30 &  91.37 & 92.65 & 91.56  \\
    Supervised + regularization  & \textbf{90.34} &  \textbf{92.54} & \textbf{92.94} & \textbf{91.64}\\
    \midrule
    \multicolumn{4}{c}{ScanObjectNN accuracy} \\
    \midrule
    Supervised  & 75.95 & 85.63 & \textbf{88.24} & \textbf{78.07}\\
    Supervised + regularization  & \textbf{76.68} &\textbf{85.81} & 87.23 &  77.69\\
    \bottomrule
  \end{tabular}
    \label{supp_tab:regul_extended}
\end{table}

\begin{table}[ht!]
    \caption{\textbf{Fine-tuning of geometric regularized pre-trained models on semantic segmentation of S3DIS scenes.} For relevant architectures, regularization can improve downstream performance of supervised pre-training.}
  \centering
  \begin{tabular}{@{}lccc@{}}
    \toprule
    Pre-training Strategy & PointMLP & PCT\\
    \midrule
    \multicolumn{3}{c}{S3DIS mIoU} \\
    \midrule
    Supervised  & 55.7 &  50.7   \\
    Supervised + regularization  & \textbf{56.74} &  \textbf{51} \\
    \bottomrule
  \end{tabular}

    \label{supp_tab:regul_extended_s3dis}
\end{table}

\begin{table}[ht!]
  \centering
  \caption{\textbf{Accuracy evaluation of geometric regularization on different layers for DGCNN.} Using only the first two layers of DGCNN provides the best features for fine-tuning on ModelNet40 classification task.}
  \begin{tabular}{@{}lccccc@{}}
    \toprule
     Layer regularized & conv1 & conv2 & conv3 & conv4 & Output layer \\
    \midrule
    Supervised + layer-regularization & 92.65 & \textbf{93.34} & 93.18 & 92.9 & 93.06\\
    \bottomrule
  \end{tabular}
  \label{supp_tab:ablation}
\end{table}

\section{Regularization}
\label{supp_sec:regul}

Our geometric regularization technique proved beneficial in enhancing the performance of supervised pre-training for the DGCNN architecture, as reported in Table 3 of the main document. Expanding upon this, Tables \ref{supp_tab:regul_extended} and \ref{supp_tab:regul_extended_s3dis} illustrate the effectiveness of this method for additional architectures across various tasks. The observed improvements in most settings indicate the versatility of our regularization method in augmenting pre-training effectiveness.

Particularly, for architectures like PointMLP, which already show a preference for supervised over contrastive pre-training, geometric regularization further elevates performance in challenging scenarios, such as S3DIS scene segmentation, where supervised pre-training alone may fall short.

Through empirical testing to determine the optimal layers for regularization, we discovered that targeting early layers for regularization yields better outcomes than applying it to the encoder's output layer alone. This finding, presented in Table \ref{supp_tab:ablation}, suggests future research needs to carefully consider which layers to regularize, as early layers appear more beneficial for this purpose, potentially guiding more effective regularization strategies in 3D model pre-training.

\end{document}